\documentclass[11pt]{article}

\usepackage[preprint]{acl}

\usepackage{times}
\usepackage{latexsym}

\usepackage[T1]{fontenc}

\usepackage[utf8]{inputenc}

\usepackage{microtype}

\usepackage{inconsolata}

\usepackage{graphicx}

\usepackage[table]{xcolor}
\usepackage{multirow}
\usepackage{booktabs}
\usepackage{caption}
\usepackage[most]{tcolorbox}
\usepackage{bm}
\usepackage{mathtools}
\usepackage{arydshln}
\usepackage{amssymb}

\usepackage{listings}
\usepackage{tcolorbox}
\tcbuselibrary{listings, skins}

\definecolor{darkgreen}{rgb}{0.0, 0.5, 0.0}
\colorlet{codeblue}{blue!10}
\colorlet{codepink}{red!10}

\newcommand{\bluebg}{%
  \rlap{%
    \color{codeblue}%
    \hspace{-3.2em}
    \rule[-0.35em]{\dimexpr\linewidth}{1.2em}
  }%
}

\newcommand{\pinkbg}{%
  \rlap{%
    \color{codepink}%
    \hspace{-3.2em}
    \rule[-0.35em]{\dimexpr\linewidth}{1.2em}
  }%
}

\newcommand{\name}{MPI}
\newcommand{\Name}{Manifold Power Iteration}

\title{Redesign Mixture-of-Experts Routers with Manifold Power Iteration}

\author{
    \textbf{Songhao Wu\textsuperscript{1}}
    \quad \textbf{Ang Lv\textsuperscript{1}}
    \quad \textbf{Ruobing Xie\textsuperscript{2}\thanks{Correspondence to: Ruobing Xie, Yankai Lin.}}
    \quad \textbf{Yankai Lin\textsuperscript{1}\footnotemark[1]} \\
    \textsuperscript{1} Gaoling School of Artificial Intelligence, Renmin University of China \smallskip \\
    \textsuperscript{2} Large Language Model Department, Tencent \smallskip \\
    \texttt{\{songhaowu, anglv, yankailin\}@ruc.edu.cn} \quad \texttt{xrbsnowing@163.com} \\
}

\usepackage{amsmath,amsfonts,bm}

\def\eqref#1{equation~\ref{#1}}

\def\1{\bm{1}}

\def\vr{{\bm{r}}}

\def\vw{{\bm{w}}}
\def\vx{{\bm{x}}}

\def\mE{{\bm{E}}}

\def\mG{{\bm{G}}}

\def\mI{{\bm{I}}}

\def\mM{{\bm{M}}}

\def\mR{{\bm{R}}}

\def\mW{{\bm{W}}}

\DeclareMathAlphabet{\mathsfit}{\encodingdefault}{\sfdefault}{m}{sl}
\SetMathAlphabet{\mathsfit}{bold}{\encodingdefault}{\sfdefault}{bx}{n}

\begin{document}
\maketitle
\begin{abstract}

Router is the cornerstone component to the Mixture-of-Experts models.
Serving as expert proxies, the rows of the router matrix compute their similarity to the MoE inputs to determine which subset of experts is activated.
Ideally, each router row is designed to encode the  expert matrix into this representative vector, such that its dot-product with token can better reflect token-expert affinity.
However, there exists no design principles to enforce this condensation.
In this paper, we propose to align each router row with the principal singular direction of the associated expert, as this direction provides the most expressive mathematical description of a matrix.
Based on this principle, we propose a router redesign with Manifold Power Iteration (MPI).
Specifically, it introduces a ``Power-then-Retract'' paradigm, where a power iteration step is performed on the router weights, followed by a retraction to impose a norm constraint to ensure both efficiency and stability.
Theoretically, we show that MPI drives router rows to converge toward the principal singular directions of associated experts.
Empirically, we pretrain MoE model across scales from 1B to 11B parameters to confirm that this alignment facilitates more effective MoE models.

\end{abstract}

\section{Introduction}
Mixture-of-Experts (MoE, \citealp{olmoe,gptoss,deepseekv4,glm5}) stands as a pivotal model architecture in LLMs to scale model capacity with a constrained computational budget.
Specifically, it replaces standard Transformer Feed Forward Networks (FFNs) with an ensemble of expert modules, using a router to select experts per token for sparse activation.
MoE enables more efficient training given the same computation budget, paving the way for LLM training with trillions of parameters~\cite{deepseekv4,k2}.

At the heart of MoE lies the router, which is typically parameterized as a linear matrix. 
For each input token, the router computes similarity scores against the matrix rows and dispatches the token to the experts corresponding to the top-scoring rows. 
While this design is straightforward and has long been accepted as a matter of course, we challenge this conventional wisdom.
Ideally, each individual row in MoE router matrix should faithfully reflect the expert's intrinsic features.
The router matrix can thus better ground the identity of each expert,
allowing token–router affinity to serve as a precise proxy for token–expert assignment.
However, there lacks a constraint in MoE router to enforce the encoding of expert features into router rows of limited expressivity.
This absence may lead to suboptimal router design, compromising both training convergence and competence of MoE models.

We propose to align each router row with the principal singular direction of its corresponding expert's weight matrix. 
This choice is grounded in a linear algebraic intuition: the principal singular direction preserves the highest density of information within a matrix~\cite{matrixcomputation,power}, making it the optimal compressed representation to characterize that matrix. 
Since each expert module is parameterized as weight matrices, encoding it into a single router vector is exactly the task of capturing its most informative direction.
To avoid the prohibitive cost of exact singular value decomposition (SVD), we leverage power iteration~\cite{power} as a lightweight alternative to obtain this principal direction online.
Specifically, the power iteration scheme uses only standard matrix-vector products to solve for the principal singular vector, obviating the need for expensive full matrix factorization.

In practice, we perform only one single power iteration on the router weights during each training step.
After that, a retraction step is introduced to regularize the $L_2$ norm of the router weights, maintaining them at a constant scale to prevent potential explosion or collapse. 
This ``Power-then-Retract" paradigm gives our design its name: \textbf{Routers with Manifold Power-Iteration (\name)}.
We prove that this online update rule is equivalent to a steepest ascent optimization that maximize the router's projection onto the expert weight under the principle of minimal updates.
From a theoretical perspective, this confirms that each update step drives an adaptive convergence of router rows toward the principal singular direction of their associated experts.
Consequently, this imposes an explicit constraint on router optimization to encode the most dominant expert features into router vectors, which has been overlooked in conventional router designs.

We conduct extensive pretraining experiments across a wide range of MoE model scales using billions of tokens.
We contend that our router redesign with \Name\ presents a fundamental departure from conventional MoE routers and brings intrinsic improvements.
While it retains the standard interface of MoE routers, it provides a fresh perspective to rethink the interplay between routers and experts.
Empirical evaluations, scaling up to 11B parameters, show that MoE with \name\ consistently facilitates faster convergence, superior downstream performance, and improved load balancing. 
We further demonstrate that this superiority is robust to shifts in model features, stemming entirely from our principled router design.
We hope these insights shed light on the intrinsic nature of MoE router and inspire future exploration.

\section{Background: Mixture-of-Experts}
We center our discussion on MoE-based LLMs. 
The key component of an MoE is the router, which dispatches inputs to a sparse subset of the experts.
Typically, the router employs a 2D linear weight matrix \(\mR \in \mathbb{R}^{N\times D}\) to project the input \(\vx \in \mathbb{R}^D\) into gating weight vector \(\vw\) over the \(N\) experts:
\begin{equation}
    \vw = \texttt{Softmax}\left(\texttt{TopK}\left(\vx\mR^\top\right)\right).
\end{equation}
where experts with the top-$K$ largest gating weights are selected.
MoE layer output is then computed as weighted sum of the selected experts:
\begin{equation}
    \textrm{MoE}\left(\vx\right) \; = \; \sum_{k=1}^K{\vw_k \cdot \mE_k(\vx)}, 
\end{equation}
where each expert module is Gated Linear Unit~\cite{glu} with the Swish~\cite{swish} activation function:
\[
    \mE_k(\vx) = \left(\texttt{SiLU}(\vx\mW^{k}_{g}) \odot (\vx \mW^{k}_{p})\right)  \mW^{k}_{o} \;.
\]
While this router design is straightforward and sufficient in most cases, certain limitations persist at the design stage and  hinder its optimal performance.
For instance, no explicit constraint is imposed to ensure that routers can faithfully reflect the experts' intrinsic features.
For input $\vx$, its affinity with the $i$-th expert is defined as its inner product with $\mR_{[i]}$.
Ideally, $\mR_{[i]}$ should maximally preserve the geometry of the $i$-th expert weights to better act as a feature vector; however, such a constraint is absent and may result in suboptimal convergence as a consequence.
Leveraging this insight, we propose a redesign of MoE router and empirically confirm its effectiveness in the following sections.

\section{Methodology}
We first elucidate our motivation, derive framework for Manifold Power Iteration and interpret some key design principles.
We then revisit the essence of our method from optimization perspective and provide accessible insights into how it works.

\subsection{Motivation}
\label{sec:motivation}
In MoE routing, \(\mathbf{R}_{[i]}\) is designed to serve as a representative vector for the \(i\)-th expert, ensuring that its inner product with an input faithfully reflects their mutual affinity. 
Consequently, the token is routed to the experts with the highest affinity scores.
This suggests that an ideal \(\mR_{[i]}\) should be optimized to effectively encode the distinctive characteristics of the expert matrix \(\mW_*^i\) within a constrained vector space. 
From a matrix-theoretic perspective, a vector is best aligned with a matrix’s principal singular directions to capture its most essential traits~\cite{Eckart1936TheAO}. 
In the context of MoE, this principle dictates that a well-coupled router \(\mR_{[i]}\) should be guided toward the principal singular direction of expert weights $\mW_*^i$. 
Geometrically, this is equivalent to maximizing squared projection of \(\mathbf{R}_{[i]}\) onto the row space spanned by \(\mathbf{W}_*^i\), which is given by:
\begin{equation}
    \max_{\mR_{[i]}} \quad {\bm{\phi}(\mW_{*}^i, \, \mR_{[i]}) = \frac{\|\mR_{[i]}\mW_*^i\|_2^2}{\|\mR_{[i]}\|_2^2}} 
    \label{eq:objective} 
\end{equation}
where $\bm{\phi}(\cdot)$ is the objective function, also known as the Rayleigh quotient with  $\mW_*^i\mW_*^{i\top}$ and $\mR_{[i]}$\footnote{
Unless specified otherwise, we substitute $\mathbf{W}_g^i$ for $\mathbf{W}_*^i \in \{\mathbf{W}_g^i, \mathbf{W}_o^i, \mathbf{W}_p^i\}$ hereafter for the sake of simplicity.
}

However, it is prohibitive to execute an exact singular value decomposition (SVD) for all expert matrices to obtain the principal singular vector at each training step.
To address the issue, we leverage power iteration onto $\mR_{[i]}$ as a lightweight alternative to SVD. 
Backed by power method theory~\cite{matrixcomputation}, it enables $\mathbf{R}_{[i]}$ to progressively track and converge toward the principal singular direction over training steps through efficient matrix-vector products.
This motivates the core design of our routers, which is built upon Power Iteration followed by a row-wise normalization.
We first outline the implementation details and defer an in-depth discussion to a later section.

\subsection{Routers with Manifold Power-Iteration}
\label{sec:definition}
\paragraph{Manifold Power-Iteration.}
Specifically, the proposed approach follows a "Power-then-Retract" paradigm, which involves (1) a power iteration step that aligns the router with the principal direction of expert weights,
followed by (2) a $L_2$ retraction step for weight containment and numerical stability.

For an arbitrary row $\mR_{[i]}$ of the router weights, we first fetch its associated expert weights $\mW_g^i$  and perform a single step of power iteration on it:
\begin{equation}
    \hat{\mR_{[i]}} \; = \; \mR_{[i]} \; \mW_g^i \; \mW_g^{i\top}.
    \label{eq:power}
\end{equation}
The cumulative execution of power iteration across training steps can induce numerical instability, causing $L_2$ norm of $\hat{\mathbf{R}}_{[i]}$ to diverge.
To counteract this divergence, we constrain the $L_2$ norm of $\hat{\mathbf{R}}_{[i]}$ to a hyperparameter $C$ after each iteration:
\begin{equation}
    \mR_{[i]}^\prime = C \cdot \frac{\hat{\mR_{[i]}}}{\|\hat{\mR_{[i]}}\|_2},
\end{equation}
while designed to avoid instability, this retraction provides additional benefits. 
Conceptually, it rectifies the potential expert bias induced by scale disparities in router norms, where an amplified norm can easily inflate gating weights and consequently overload the corresponding expert. 
Based on these two designs, the original router matrix $\mR$ is substituted with the concatenated formulation:
\[
    \mR^{\prime\top} = \big[ \, \hat{\mR_{[1]}} \mid \hat{\mR_{[2]}} \mid \cdots \mid \hat{\mR_{[N]}} \, \big],
\]
and the final gating weights \(\vw\) are recomputed as:
\begin{equation}
    \vw^\prime = \texttt{Softmax}\left(\texttt{TopK}\left(\vx\mR^{\prime\top}\right)\right).
\end{equation}
We designate this refined router $\mR^\prime$ as \textbf{Routers with Manifold Power-Iteration (MPI)}, to fully manifest the Power-then-Retract paradigm.

Figure~\ref{fig:code} provides a Pytorch-style pseudo-code to help understand our implementation.
\begin{figure}[h]
\begin{lstlisting}[language={Python}]
from torch.nn.functional import normalize
from megablocks.layers.moe import MoE
class MoE_MPI(MoE):
    def foward(self, x, C_prime = 1):
        # R: [N, D], wg: [N, D, d]
        # PowerIter: ([N, 1, D] @ [N, D, d]) 
        #   @ [N, d, D] -> [N, D]
        (*\bluebg*) R_hat = (self.R.unsqueeze(1) @ wg.
        (*\bluebg*)    transpose(1, 2) @ wg).squeeze()
        # Retracition: [N, D] -> [N, D] 
        (*\pinkbg*) R_prime = normalize(R_hat, p=2, dim=-1)
        (*\pinkbg*) C = C_prime * (N ** -0.5)
        (*\pinkbg*) logits = C * (x @ R_prime.T)
        s_prime = logits.softmax(dim=-1)
        w_prime, _ = torch.topk(s_prime, dim=-1)
        return self.experts(x, s_prime, w_prime)        
\end{lstlisting}
\caption{Pseudo code for Manifold Power-Iteration.}
\label{fig:code}
\end{figure}

\paragraph{Design Principle.} We also establish a principle to guide the configuration of $C$.
To this end, we introduce an assumption that routing logits should be bounded at a constant scale to to prevent explosion, inspired by insights in~\citep{scion}:
\[
    {\| \, \vx\mR^{\prime\top} \, \|_{\infty}} \; \sim \; O(1),
\]
Given a scale-invariant $\mathbf{x}$, this upper bound inherently implies that $C \sim \Theta(\frac{1}{\sqrt{N}})$ with respect to $N$.
This is evidenced by the following derivation:
\begin{gather}
    \| \, \vx\mR^{\prime\top} \, \|_{\infty} \le \, \sqrt{\sum_{i=1}^N{(\vx\mR_{[i]}^\top)^2}} \; \sim \; O(C\sqrt{N}), \nonumber \\
    \texttt{where} \quad \vx\mR_{[i]}^\top \; \sim \; O(C) \quad \text{for each expert}.
\end{gather}
Therefore, to enforce the $O(1)$ ceiling and decouple the scaling effect from the expert count $N$, we introduce a redefinition: $C \coloneqq \frac{C^\prime}{\sqrt{N}}$, where $C^\prime$ is a scale-invariant global hyperparameter.

\subsection{From Maximum Projection Constraints to Manifold Power-Iteration}
\label{sec:update}
Section~\ref{sec:motivation} provides an intuitive motivation explaining the introduction of Power-Iteration into our router design.
This section extends the maximum projection objective in Eq.~\ref{eq:objective} to align with our formulation, which can be expressed as:
\begin{gather}
    \max_{\Delta \vr} \quad \bm{\Phi}(\mW_g, \, \mR_{[i]}^\prime + \Delta \vr) \\
    \mathrm{\textbf{s.t.}} \,\; \|\mR_{[i]}^\prime\|_2 = \|\mR_{[i]}^\prime + \Delta \vr\|_2 = C, \;\; \|\Delta \vr\|_2 \le \eta, \nonumber
\end{gather}
since the normalization ensures a constant denominator, the original objective reduces to:
\[
    \bm{\Phi}(\mW_{*}^i, \, \mR_{[i]}^\prime) = \|\mR_{[i]}^\prime\mW_*^i\|_2^2 = \mR_{[i]}^\prime\mW_*^i\mW_*^{i\top}\mR_{[i]}^{\prime\top},
\]
$\Delta \vr$ represents the update, and $\eta$ constrains the update within a small bounded region. 
We impose norm constraints on both $\mR_{[i]}^\prime$ and $\mR_{[i]}^\prime + \Delta \vr$ .
To analyze this optimization landscape, we consider a first-order Taylor approximation of the objective:
\begin{equation*}
    \bm{\Phi}(\mW_g, \, \mR_{[i]}^\prime + \Delta \vr) \; = \; \bm{\Phi}(\mW_g, \, \mR_{[i]}^\prime)  \, + \, \langle \mG, \, \Delta \vr \rangle,
\end{equation*}
where $\mG = 2 \, \mR_{[i]}^\prime \mW_g \mW_g^\top$ represents the gradient of $\mR_{[i]}$.
The Taylor approximation reduces the objective to maximizing the inner product $\langle \mG, \, \Delta \vr \rangle$.
To satisfy the norm constraint and ensure that the updated router $\mR_{[i]}^\prime$ remains on the spherical manifold, we project the gradient $\mG$ onto the tangent space of the sphere. 
Defining $\mM \coloneqq \mW_g\mW_g^\top$ to simplify notation, and setting $C = 1$ without loss of generality, the gradient ascent update $\Delta \vr_{g}$ on the manifold is formulated as:
\begin{align}
    \Delta \vr_{g} &= \eta \,\, \mG \, \left( \mI - \frac{\mR_{[i]}^{\prime\top}\mR_{[i]}^\prime}{\mR_{[i]}^\prime\mR_{[i]}^{\prime\top}}  \right) \nonumber \\
        &= \eta \left( \mR_{[i]}^\prime\mM - \mR_{[i]}^\prime \left(\mR_{[i]}^\prime \mM\mR_{[i]}^{\prime\top}\right) \right), 
\end{align}
where the scaling constants are absorbed into $\eta$.
We also derive an approximation for the exact update $\Delta \vr_{M}$ introduced by Manifold Power-Iteration:
\begin{equation} 
   \Delta \vr_{M} \approx \frac{1}{\mR_{[i]}^\prime \mM \mR_{[i]}^{\prime\top}} {\left( \mR_{[i]}^\prime \mM - \mR_{[i]}^\prime(\mR_{[i]}^\prime \mM \mR_{[i]}^{\prime\top}) \right)}.
   \label{eq:10}
\end{equation}
By comparing steepest ascent ($\Delta \vr_{g}$) with our router update ($\Delta \vr_{M}$), we observe a striking structural alignment.
In this light, our proposed router design constitutes an optimization tailored for maximum projection constraints with an adaptive step-size.
Specifically, it drives a steady convergence of the router weights toward the principal singular vector, with the step size decreasing and updates becoming more careful as $\mR_{[i]}^\prime$ are mostly aligned with the principal direction of $\mW_g^i$ (i.e. $\Delta \vr_{M}$ is moderated because the denominator $\mR_{[i]}^\prime \mM \mR_{[i]}^{\prime\top}$ in Eq.~\ref{eq:10} increases), and vice versa.

The update can also be interpreted from an SVD perspective. After sufficiently many training steps (or power iterations), the term $\mR_{[i]}^\prime \mM$ in Eq.~\ref{eq:10} approaches the dominant singular vector of $\mW_g$.
At that stage, the scalar quantity $\mR_{[i]}^\prime \mM \mR_{[i]}^{\prime\top}$ corresponds to the $L_2$ norm when feeding $\mR_{[i]}^\prime$ into $\mW_g$, yielding a scalar that scales $\mR_{[i]}^\prime$. 
The subtraction term in Eq.~\ref{eq:10} therefore derives an update direction that points toward the residual mismatch between $\mR_{[i]}$ and the dominant singular vector. 
Applying the update progressively rotates $\mR_{[i]}$ toward the principal singular subspace of $\mW_g$.

These interpretations help explain why the proposed method effectively optimizes the router to encode the most informative expert features.
We provide supplementary derivation in Appendix~\ref{apx:math} and hope this perspective can inspire readers.

\begin{figure}[t] 
    \centering
    \begin{minipage}{\columnwidth}
        \centering
        \includegraphics[width=\columnwidth]{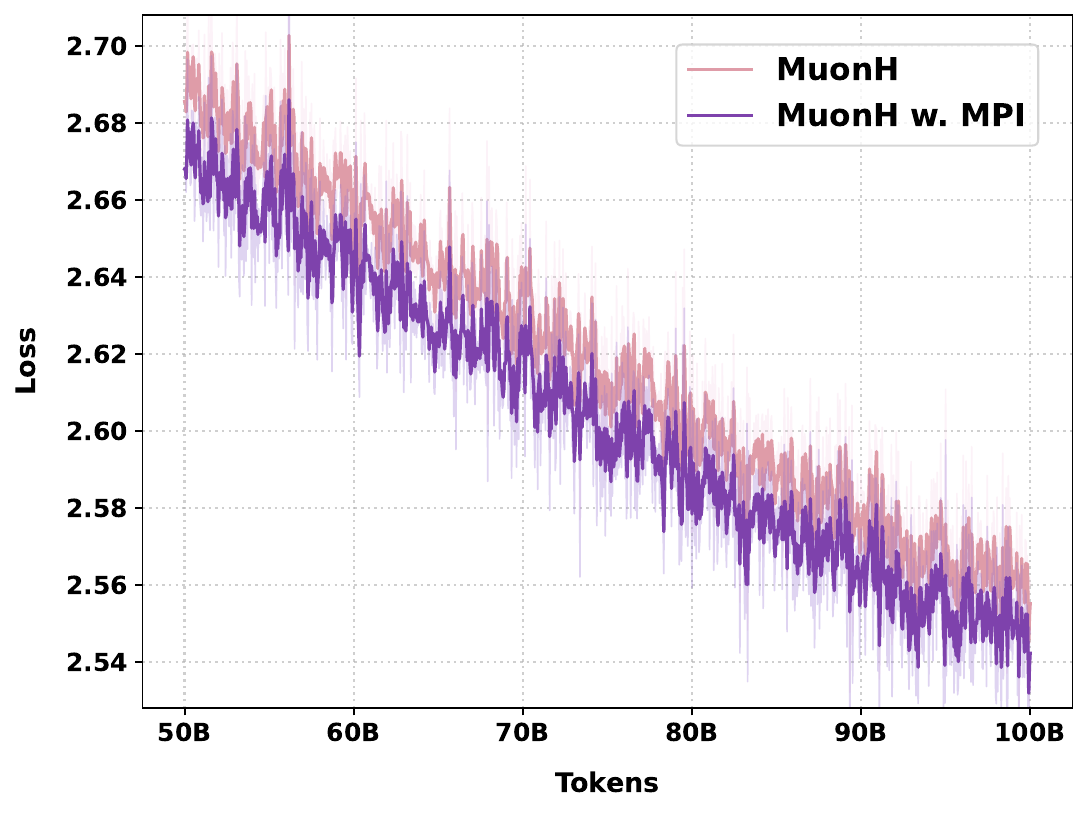}
        \caption{Convergence comparisons for MoE with \name, exemplified by \textcolor{gray}{MuonH-1B}. Our router design achieves a 0.013 reduction in pretraining loss. Similar observations for other optimizers are provided in the Appendix.}
        \label{fig:loss-1b-comp}
    \end{minipage}
    \vspace{1em} 
    \begin{minipage}{\columnwidth}
        \centering
        \small
        \vspace{1em}
        \resizebox{0.96\columnwidth}{!}{%
            \begin{tabular}{l|cccc} 
                \toprule
                 & AdamW & AdamH & Muon & MuonH \\
                \midrule
                MoE & 42.26 & 42.59 & 43.01 & 42.78 \\
                \cmidrule(lr){1-1} \cmidrule(lr){2-5}
                \textbf{+} \name & 43.56 & 43.93 & 43.55 & 43.98 \\
                \bottomrule
            \end{tabular}%
        }
        \captionof{table}{Downstream performance (average accuracy across \textbf{25 benchmarks}).
        \name~consistently improves downstream performance across different optimizers.
        Detailed task-specific results are provided in Table~\ref{tab:full}. In the remainder of this paper, unless otherwise specified, we only report the average results across the 25 tasks.
        }
        \vspace{1em}
        \label{tab:opt-agnostic}
    \end{minipage}
\end{figure}

\begin{figure*}[t]
    \centering
    \includegraphics[width=0.92\linewidth]{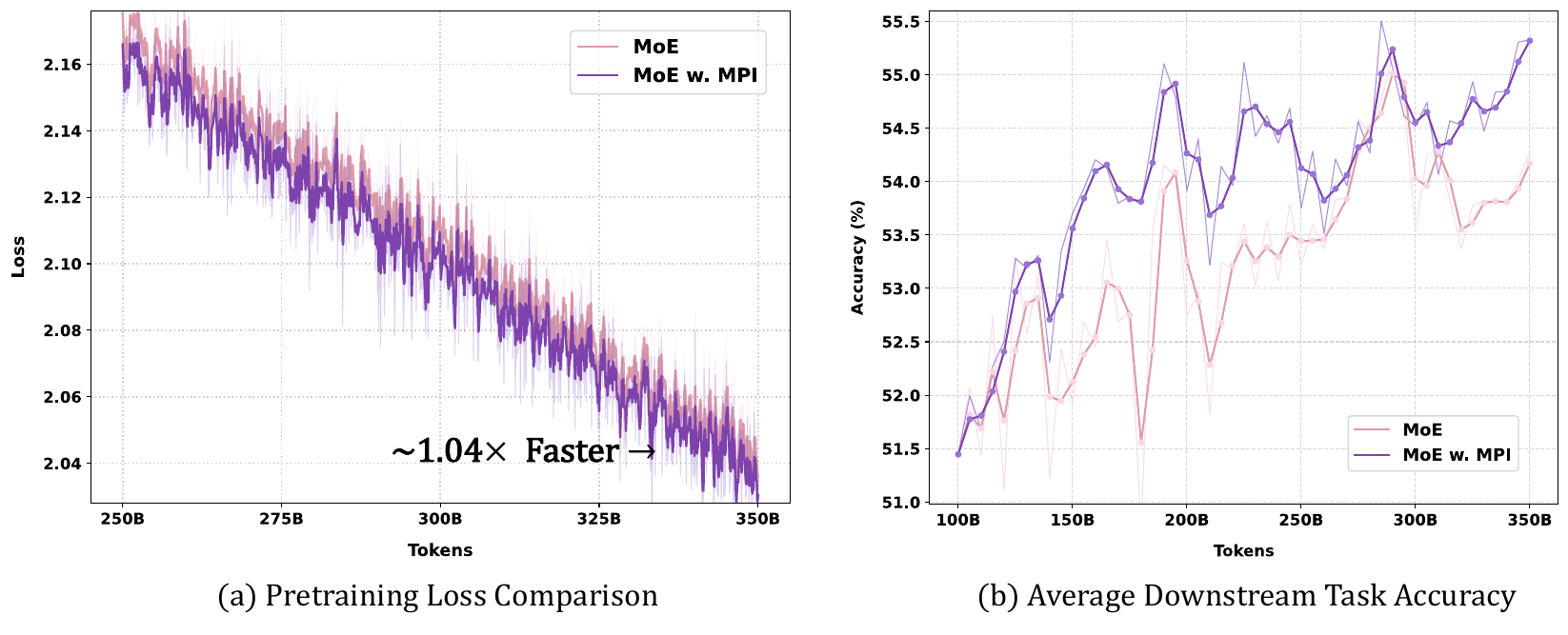}
    \caption{Convergence and Downstream Performance Comparison. \Name\ facilitates faster convergence and superior downstream task performance throughout the entire course of 11B MoE pretraining.}
    \label{fig:loss-task}
    \vspace{-2mm}
\end{figure*}

\section{Experiment}
\subsection{\name\ is an Optimizer-Agnostic Design}
We first pretrain 1B MoE models using different optimizers, guided by two primary motivations:

(1) To substantiate that \name\ is an intrinsic improvement to router design, which remains agnostic to shifts in model features across optimizers

(2) To provide a foundational analysis to justify our setup for the large-scale experiments.

Specifically, we pretrain these 1B models with AdamW~\cite{adamw} and Muon~\cite{muon}, alongside their Hyperball Opimization~\cite{hyperball} variants, AdamH and MuonH.\footnote{For readers unfamiliar with these advanced optimizers, please refer to the related work section (Section~\ref{sec:related}).}
Detailed model and optimizer configurations are provided in Appendix~\ref{apx:details}.

We pretrain the baselines on 100B tokens and analyze the resulting convergence and downstream performance.
Figure~\ref{fig:loss-1b-comp} plots the convergence comparisons, using MuonH as representative.
Table~\ref{tab:opt-agnostic} reports the average downstream performance over a task suite of 25 benchmarks.
Crucially, MoE with \name\ achieves both accelerated convergence and improved downstream performance across all optimizer setups at the 1B scale. 
Motivated by this, we scale our experiments to confirm these benefits at larger capacities. 
We select MuonH for the large-scale experiments owing to 
(1) its  hyperparameter transferability, 
and (2) its optimal convergence performance among all 1B MoE baselines.

\subsection{Comparative Analysis with vanilla MoE}
\label{sec:main-exp}
We pretrain MoE with \name\ at two larger scales: \textbf{3B} and \textbf{11B}.
All models are pretrained on 350B tokens sampled from FineWeb-Edu dataset~\cite{fineweb-edu},
with 1B tokens reserved to serve as the validation set.
We then midtrain the models on 100B tokens from~\citealp{olmo3}.
Full architecture hyperparameters, training configurations are available in Appendix~\ref{apx:details}.
\begin{table}[h]
    \centering
    \small
    \resizebox{0.42\textwidth}{!}{
        \begin{tabular}{lccc}
            \toprule
             \textbf{PPL~($\downarrow$)} & \textbf{Validation} & \textbf{Math} & \textbf{Code} \\
            \midrule
             MoE \textcolor{gray}{3B} & 0.764 & 1.688 & 1.376 \\
             MoE \textit{w.} \name\ & \textbf{0.754} & \textbf{1.581} & \textbf{1.296} \\
             \cmidrule(lr{1pt}){1-1} \cmidrule(lr{1pt}){2-4}
             MoE \textcolor{gray}{11B} & 0.728 & 1.852 & 1.263 \\
             MoE \textit{w.} \name\ & \textbf{0.723} & \textbf{1.581 }& \textbf{1.259 }\\
             \bottomrule
        \end{tabular}
    }
    \caption{Perplexity in bits per byte for MoE with \name.}
    \label{tab:ppl}
\end{table}
We forgo comparisons with other baselines since our design conforms to the standard router form and is theoretically orthogonal to these studies.
Section~\ref{sec:compatibility} explores this compatibility to provide an investigation.

\paragraph{Convergence and Performance.}
We conduct pretraining on two scales and confirm that MoE with \name\ achieves faster convergence and improved downstream performance.
Figure~\ref{fig:loss-task} (a) and (b) present a comparison of pretraining loss and downstream performance evolution for 11B MoE.

We observe that MoE with \name\ leads to more effective training and maintains this loss advantage throughout.
We also report perplexity comparison evaluated on both validation set and held-out Math and Code sets from Olmo 3~\cite{olmo3}.
As shown in Table~\ref{tab:ppl}, the advantage for language modeling remains consistent across all domains.
\begin{table*}[t]
    \small
    \centering 
    \resizebox{0.92\textwidth}{!}{
    \begin{tabular}{lccccccccc}
        \toprule
        Task\; ($\rightarrow$)  & ARC-C & MMLU & TriviaQA & NaturalQs & BBH & GSM8K & MBPP & AVG. \\
        \cmidrule(lr{1pt}){1-1} \cmidrule(lr{1pt}){2-3} \cmidrule(lr{1pt}){4-5} \cmidrule(lr{1pt}){6-6} \cmidrule(lr{1pt}){7-8} \cmidrule(lr{1pt}){9-9}
        Setup  ($\rightarrow$) & \footnotesize{5-shot} & \footnotesize{5-shot} & \footnotesize{5-shot} & \footnotesize{5-shot} & \footnotesize{3-shot CoT} & \footnotesize{8-shot CoT} & \footnotesize{3-shot}  \\
        \midrule
        MoE \textcolor{gray}{3B} & 55.91 & 47.01 & 45.78 & 17.87 & 29.53 & 16.22 & 42.25 & 36.37 \\
        \cmidrule(lr{1pt}){1-9} 
        \quad \textit{w.} \name\ & 58.96 & 48.83 & 46.52 & 20.13 & 30.99 & 20.92 & 44.54 & 38.70 \\ 
        \midrule
        MoE \textcolor{gray}{11B} & 61.54 & 50.00 & 55.41 & 25.30 & 31.17 & 17.89 & 45.12 & 40.92 \\
        \cmidrule(lr{1pt}){1-9}
        \quad \textit{w.} \name\ & 62.24 & 50.93 & 56.89 & 25.36 & 31.45 & 27.60 & 44.87 & 42.76 \\ 
        \bottomrule 
    \end{tabular}
    } 
    \caption{Performance of MoE with \Name\ on challenging benchmarks at both 3B and 11B scales.}
    \label{tab:midtrain}
\end{table*}

We evaluate downstream tasks to confirm that the loss reduction manifests as superior model competence.
Specifically, we use a suite of 9 core tasks to monitor and Figure~\ref{fig:loss-task} (b) plots the evolution of average accuracy throughout pretraining.
We observe that \name\ maintains the advantage on downstream tasks throughout pretraining.
Furthermore, we extend our evaluation to more challenge tasks after mid-training,
including benchmarks across knowledge-intensive QA~\cite{arc,mmlu}, reading comprehension~\cite{triviaqa,nq}, language understanding and reasoning~\cite{bbh}, math skills and code generation~\cite{gsm8k,mbpp}.
As summarized in Table~\ref{tab:midtrain}, MoE with \name\ delivers consistent performance gain, which further validates the effectiveness of our router design.
Our evaluation setups are available in Appendix~\ref{apx:eval}.

\begin{figure}[t]
\vspace{-2mm}
    \centering
    \includegraphics[width=0.9\linewidth]{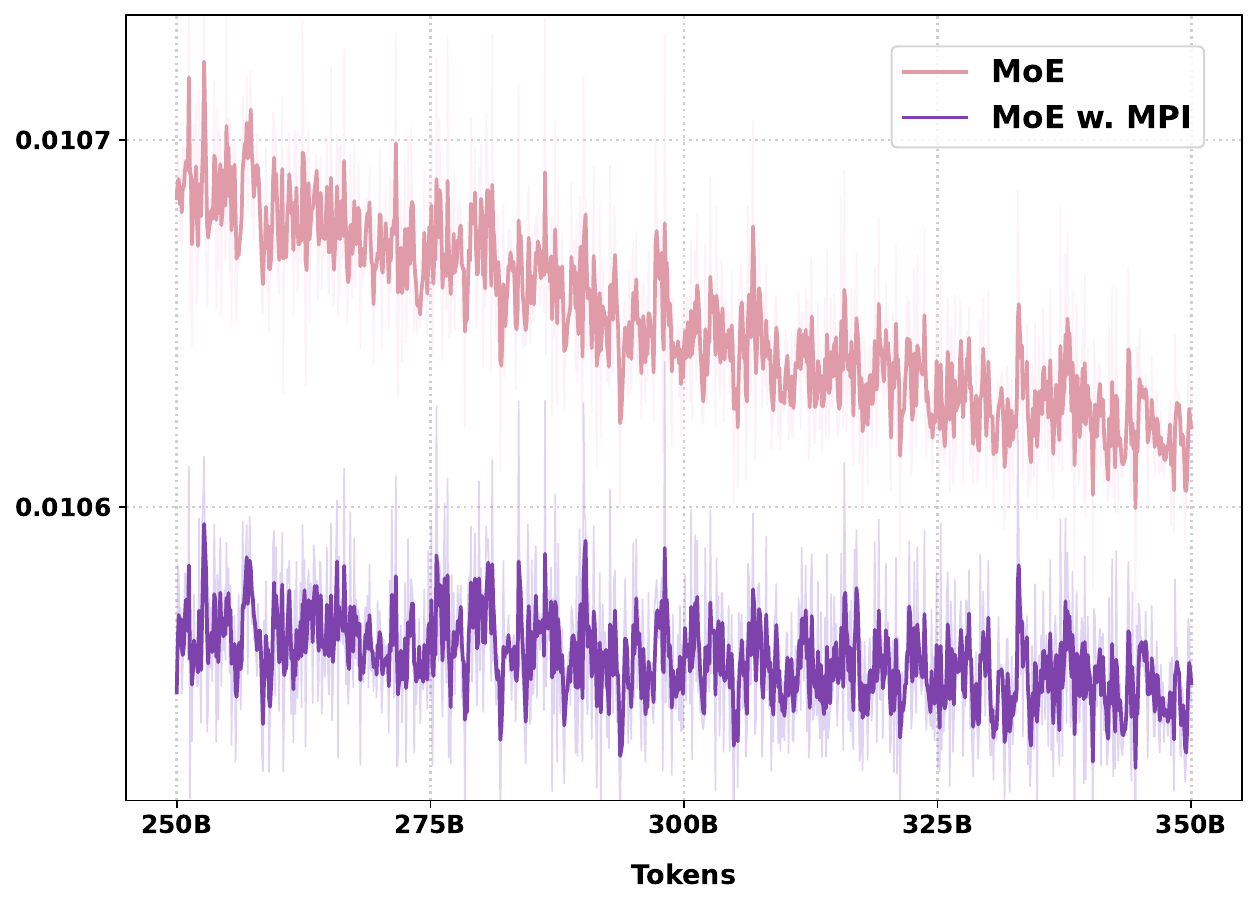}
    \caption{Load balancing loss for 3B MoE with \name.}
    \label{fig:bal-loss}
\end{figure}

\paragraph{Load Balancing.}
As shown in Figure~\ref{fig:bal-loss}, a noticeable decrease in balance loss is observed for MoE with \name\ during pretraining.
Specifically, this loss drops sharply during the early stages and remains at a low level thereafter.
We suspect that this reduction might be an artifact of router retraction.
Therefore, we report $\text{MaxVio}$ on validation set as a more accurate reflection of load balance.

Table~\ref{tab:maxvio} reports both $\mathrm{MaxVio}_{Batch}$ and $\mathrm{MaxVio}_{Global}$ of different models.
The reported $\mathrm{MaxVio}$ confirms that \name\ is compatible with the load balancing loss and achieves a more equitable load distribution than the vanilla MoE as an unexpected bonus.
Tentatively, we attribute the improved balance to our retraction design, and leave a deeper investigation into it to future work.

\paragraph{Efficiency Analysis.}
We provide a breakdown efficiency analysis into MoE with \name\ to confirm its practicality in large-scale MoE pretraining.

(1) Our router design introduces negligible overhead with respect to training efficiency.
In our 11B pretraining experiments, vanilla MoE sustains a throughput of 34.97 billion tokens per day, while \name\ incurs a mere slowdown of 0.2\%.
Intuitively, the computational cost exerted by \name\ does not exceed that of $N$ extra tokens, which is a negligible fraction of the total tokens per batch.
Our \name\ design introduces zero communication overhead and avoids conflicts with standard training frameworks.

(2) At inference time, the router weights can be pre-computed with power iteration as the model loads.
Therefore, our design incurs zero inference overhead and maintains compatible with standard inference engines out-of-the-box.

Taken together, we believe in scalability of \name\ toward larger-scale MoE training and deployment.

\begin{table}[t]
    \centering
    \small
    \resizebox{0.78\columnwidth}{!}{%
        \begin{tabular}{lclc}
            \toprule
            \multicolumn{2}{c}{$\mathrm{MaxVio_{Batch}} \,\; (\downarrow)$} & \multicolumn{2}{c}{$\mathrm{MaxVio_{Global}} \,\; (\downarrow)$} \\
            \cmidrule(lr{1pt}){1-2} \cmidrule(lr{1pt}){3-4} 
            MoE & \textit{w.} \name\ & MoE & \textit{w.} \name\ \\
            \midrule
            1.133 & 1.024 & 0.964 & 0.711 \\
            \bottomrule
        \end{tabular}%
    }
    \caption{$\mathrm{MaxVio}$ comparisons for 3B MoE with \name.}
    \label{tab:maxvio}
\end{table}

\section{Method Analysis}
\subsection{Enhanced Router-Expert Alignment Along the Principle Singular direction}
We perform a post-hoc parameter analysis to verify that our design better aligns  router rows  with the principal singular vector of the associated experts.
Following Section~\ref{sec:definition}, we report the projection of $\mR_{[i]}^\prime$ onto $\mW_g^i$ as the quantitative metric:
\begin{equation}
    \lambda = \frac{\| \mR_{[i]}^\prime \mW_g^i\|_2}{\|\mR_{[i]}^\prime\|_2 \; \|\mW_g^i\|_2},
\end{equation}
where $\lambda$ is normalized by the spectral norm to constrain within $[0,1]$.
Table~\ref{tab:lambda} compares the $\lambda$ distributions, where the average values across experts per layer are reported.
Compared to vanilla MoE, MoE with \name\ achieves a tighter couple of router vectors with the principal directions of expert weights, manifest as a prominently higher $\lambda$.

\begin{table*}[t]
    \centering
    \small
    \resizebox{0.92\textwidth}{!}{
        \begin{tabular}{lcccccccccccc}
            \toprule
            Layer & 1 & 2 & 3 & 4 & 5 & 6  & 7 & 8 & 9 & 10 & 11 & 12   \\
            \midrule
            MoE & 0.37 & 0.31 & 0.31 & 0.28 & 0.28 & 0.25 & 0.24 & 0.23 & 0.24 & 0.24 & 0.22 & 0.26 \\
            \cmidrule(lr{1pt}){1-1} \cmidrule(lr{1pt}){2-13}
            \textit{\quad w.} \name\ & 0.67 & 0.66 & 0.69 & 0.70 & 0.68 & 0.67 & 0.64 & 0.63 & 0.62 & 0.62 & 0.62 & 0.69 \\
            \bottomrule
        \end{tabular}
    }
    \caption{Comparison of $\lambda$ distributions. Router with \Name\ exhibits an enhanced alignment between $\mR_{[i]}^\prime$ and the principal singular direction of expert weights, manifested by significantly larger $\lambda$ values.}
    \label{tab:lambda}
    \vspace{-2mm}
\end{table*}

The analysis in Section~\ref{sec:update} explains why a single power iteration suffices to achieve router–expert alignment along the principal singular direction.
Readers may wonder whether additional iterations could further enhance through a tighter alignment.
To investigate this, we increase the iteration count to 10 to ensure full convergence.
We observe that this more precise estimation results in a 5\% lower throughput, 
and provides no further convergence advantage or downstream performance improvement (with a pre-training loss increase of 0.002 to 0.003 and a downstream drop of 1.39 percentage points).
In our view, aggressive alignment  disrupt the stability of router optimization, making a single power iteration a more robust and efficient choice.

\subsection{Ablation Studies}

We conduct ablation studies to validate the design choices of routers with Manifold Power-Iteration. 

\subsubsection{Impact of the Key Design Choices}
We pretrain ablated 3B models on 200B tokens to validate the effectiveness of the two core designs: 
(1) Power Iteration and (2) Router Retraction.

\begin{figure}
    \centering
    \includegraphics[width=0.92\linewidth]{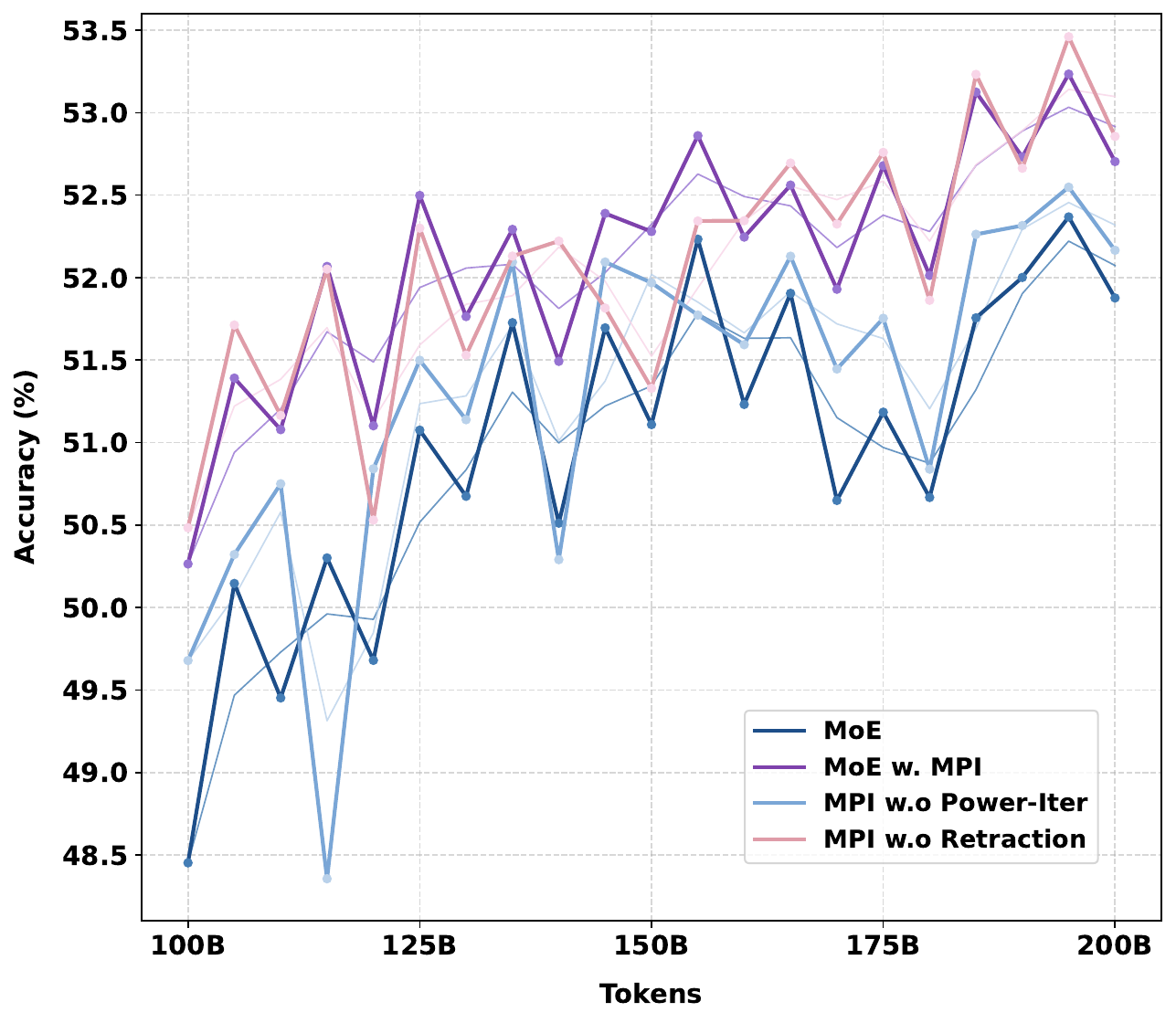}
    \caption{
    Ablation studies for the key design choices: (1) Power Iteration and (2) Router Retraction. 
    We observe pretraining collapses without Router Retraction when using AdamW and Muon, showcasing that Router Retraction is critical for maintaining training stability, especially for optimizers that lack weight constraints.
    }
    \label{fig:ablation}
\end{figure}

\paragraph{Ablation on Power Iteration Design.}
We introduce a baseline that solely performs row-wise normalization on router weights $\mR$. 
This replaces the original $\mR_{[i]}^{\prime}$ with $\mR_{[i]}^{\mathrm{np}}$, which is defined as:
\[
    \mR_{[i]}^{{np}} \, = \, C \cdot \frac{\mR_{[i]}}{\|\mR_{[i]}\|_2}.
\]
As shown in Figure~\ref{fig:ablation}, this ablated variant underperforms our routers with \name, achieving nearly identical performance to the vanilla MoE.
This confirms that our improvements cannot be attributed to router weights retraction.
However, we observe that it exhibits a similar balance loss distribution to that of our \name.
We leave it to future work to investigate whether this normalization can lead to improved load balancing as a side benefit.

\paragraph{Router Retraction Enables Stable Training.}
To resolve the instability caused by power iteration,
we adopt router retraction to mitigate the risk of $L_2$-Norm explosion or collapse.
We replace $\mR_{[i]}^\prime$ with $\hat{\mR_{[i]}}$ and first conduct ablation on 1B models.

Specifically, we observe loss spikes and abnormal gradients for 1B baselines pretrained with AdamW and Muon. 
In the absence of router retraction, the power iteration destabilizes pretraining and leads to suboptimal model convergence. 
While hyperball optimization can relieve this instability, it impose no constaint on the spectral norm of expert matrices, risking $L_2$ norm collapse as $N$ increases. 
Although the ablated variant remains competitive on downstream tasks, we observe its elevation in pretraining loss of 0.003.
Combining our empircial observations and analysis, we strongly advocate for this retraction design.

\subsubsection{Sensitivity Analysis of Constant $C$}
We benchmark small-scale MoE models with 256 experts, conducting a hyperparameter search over over $C^\prime \in \{1,2,4,8\} $.
Each model variant is pretrained on 50B tokens and optimized with MuonH.
Table~\ref{tab:choice-of-c} presents the validation perplexity (PPL) across different choices of $C^\prime$.
\begin{table}[t]
    \centering
    \small
    \resizebox{\columnwidth}{!}{
    \begin{tabular}{cccccc}
        \toprule
        $\mathbf{C^\prime}$ & \textbf{1} & \textbf{2} & \textbf{4} & \textbf{8} & \textbf{MoE}  \\
        \midrule
        \textbf{Val PPL} & 0.8896 & 0.8547 & 0.8533 & 0.8563 & 0.8884 \\
        \bottomrule
    \end{tabular}
    }%
    \caption{Validation perplexity across choices of $C^\prime$.}
    \label{tab:choice-of-c}
\end{table}

Specifically, we have the following observations:
(1) In most cases, MoE with \name\ outperforms the vanilla MoE, which demonstrate that our router design is relatively insensitive to the choice of $C^\prime$; 
(2) The optimal choice of $C^\prime$ basically aligns with the design principles we established in Section~\ref{sec:definition}.
Leveraging the Hyperball Optimization properties, we directly transfer the optimal $C^\prime$ identified in small-scale sweep into our 11B pretraining. 
As is shown in Section~\ref{sec:main-exp}, no performance collapse is observed.
We argue that this hyperparameter is mostly insensitive and transferable with the help of advanced optimizer designs.

\subsection{Expert Weight Choice for Power Iteration}
We pretrain 1B MoE baselines on 50B tokens to explore the optimal choice  among the three candidate expert weight matrices ($\mW_g$, $\mW_p$ and $\mW_o$) for power iteration.
No significant divergence in pre-training loss or downstream performance is observed for these candidates.
Therefore, we adopt $\mW_g$ as our default choice, as it holds a marginal advantage across all candidates in current experimental setup.
We leave it to future work to explore the potential of expert matrices combinations.

\section{Compatibility of Manifold Power-\- Iteration with other Router Designs}
\label{sec:compatibility}
Routers with \name\ preserve the gating weights computation and modify only the router weights. 
Conceptually, this refinement is orthogonal to most alternative router designs.
We pretrain 1B baselines on 50B tokens to explore this compatibility.

\paragraph{Auxiliary loss for MoE.}
In standard MoE practices, auxiliary losses are designed to regularize routing to address specific issues (load balance, expert specialization etc.).
Section~\ref{sec:main-exp} confirms the compatibility of our router design with load balancing loss.
We further integrate our method with router z-loss using a coefficient of 0.001~\cite{zloss}. 
Our small-scale trials exhibit no loss or gradient anomalies, and the z-loss variant yields a 0.68-point improvement in downstream tasks, further confirming its compatibility.

\paragraph{Alternative of Activation functions.}
By default, we adopt $\operatorname{Softmax}$ as the activation function.
In this section, we also explore $\operatorname{Sigmoid}$ as an alternative.
Specifically, we fix $C=1$ without searching to align with the Frobenius norm of MuonH.
Compared with $\operatorname{Softmax}$, the pretraining loss advantage narrows, while downstream performance stills improves from 41.64 to 42.05.
We reserve thorough exploration on $\operatorname{Sigmoid}$ and other activation functions to future work.

\section{Related Work}
\label{sec:related}
We provide an overview of the optimizers used in this paper, which are well-established for model convergence acceleration.
Beyond convergence, we seek to leverage their scalability,
in the hope that our empirical insights, from model up to 11B parameters trained on 350B tokens, can be extrapolated and validated efficacy at larger scales.

We begin with an introduction of Muon~\cite{muon}, which orthogonalizes momentum with Newton-Schulz iterations to update parameters.
Recent studies have validated its effectiveness in pretraining models with up to trillions of parameters~\cite{k2, deepseekv4}.
Further analysis interprets it as a steepest descent under spectral norm, which inspires other norm-constrained optimizer designs~\cite{scion}.

More recently, a line of work proposes imposing norm constraints on both weights and updates~\cite{hyperball, sso}.
The intuition behind is that norm constraints on weights enable stable and scalable optimization, which in turn accelerates convergence across scales and allows for hyperparameter transfer without further tuning.
This paper provides a preliminary practice of these optimizers and empirically validates their effectiveness.

\section{Conclusion}
We revisited the design of MoE routers from a row-wise expert-proxy representation perspective and proposed \Name\ (MPI).
MPI is an efficient and theoretically grounded alternative to conventional router designs, and establishes a principled connection between router representations and expert parameters.
It requires only lightweight iterative updates while maintaining scalability.
Extensive experiments validates MPI across diverse architectures and training settings.
We hope this work inspires future research on mathematically principled router design and advances the understanding of the representation geometry in MoEs.



\bibliography{custom}

\clearpage
\newpage

\appendix

\section{Supplementary Derivations for Approximation in Equation~\ref{eq:10}}
\label{apx:math}
In what follows, we provide detailed derivation for weights update $\Delta \vr$ of router $\mR_{[i]}^\prime$ within \Name.
Formally, $\Delta \vr$ is given as:
\[
    {\Delta \vr}_{M} = \frac{\mR_{[i]}^\prime\mM}{\|\mR_{[i]}^\prime\mM\|_2} -  \mR_{[i]}^\prime,
\]
where $\frac{\mR_{[i]}^\prime\mM}{\|\mR_{[i]}^\prime\mM\|_2}$ denotes the updated $\mR_{[i]}^\prime$ via power iteration.
We project $\mR_{[i]}^\prime \mM$ onto the subspace spanned by $\mR_{[i]}^\prime$ and its orthogonal complement:
\[
\begin{split}
    \mR_{[i]}^\prime &{} \mM \quad = \quad \mR_{[i]}^\prime \left(\mR_{[i]}^\prime \mM\mR_{[i]}^{\prime\top}\right) \\
   & + \underbrace{\Big(\mR_{[i]}^\prime\mM - \mR_{[i]}^\prime \left(\mR_{[i]}^\prime \mM\mR_{[i]}^{\prime\top}\right)\Big)}_{\text{orthogonal to } \mR_{[i]}^\prime}.
\end{split}
\]
As the power iteration proceeds, $\mR_{[i]}^\prime$ asymptotically toward the dominant subspace, and the orthogonal component (the second term above) becomes negligible. Consequently, we can arrive at the following approximation:
\[
    \frac{\mR_{[i]}^\prime\mM}{\|\mR_{[i]}^\prime\mM\|_2} \approx \mR_{[i]}^\prime + \frac{\mR_{[i]}^\prime \mM - \mR_{[i]}^\prime(\mR_{[i]}^\prime \mM \mR_{[i]}^{\prime\top}) }{\mR_{[i]}^\prime \mM \mR_{[i]}^{\prime\top}},
\]
through a simple rearrangement of terms, we obtain
\[
    {\Delta \vr}_{M} \approx \frac{1}{\mR_{[i]}^\prime \mM \mR_{[i]}^{\prime\top}} {\left( \mR_{[i]}^\prime \mM - \mR_{[i]}^\prime(\mR_{[i]}^\prime \mM \mR_{[i]}^{\prime\top}) \right)}. 
\]
which completes the derivation of Eq.~\ref{eq:10}.

\section{Details for Pretraining Experiments}
\label{apx:details}
\subsection{Implementation Details}
Table~\ref{tab:mdhyper} summarizes the hyperparameters of the model architectures across experiments. 
To support large-scale experiments, we scale our 3B model to 11B by expanding the experts counts from 64 to 256, resulting in a sparse MoE model with 11B total parameters and 470M activated parameters.

Our training pipeline is built upon the TorchTitan framework~\citep{torchtitan}.
For Transformer components, we adopt PyTorch’s SDPA for attention ~\citep{pytorch},
and MegaBlocks~\(\mathrm{MLP}\) for efficient MoE implementation~\citep{megablocks}.
In terms of model parallelism, we adopt Fully Sharded Data Parallel~\citep{fsdp} across all pretraining experiments.

\begin{table}[h]
  \centering
  \small
  \resizebox{\columnwidth}{!}{ %
    \begin{tabular}{lccc} 
        \toprule
            \textbf{Configuration} & \textbf{1 B} & \textbf{3 B} & \textbf{11 B} \\
        \midrule
            ~\textrm{Dimension} & 1024 & 1536 & 1536 \\
            ~\textrm{Layers} & 8 & 12 & 12 \\
        \cmidrule(lr){1-4}   
            ~\textrm{Attention Heads} & 8 & 16 & 16 \\
            ~\textrm{Key-value Heads} & 8 & 8 & 8 \\
            ~\textrm{RoPE Theta~($\theta$)} & 500,000 & 500,000 & 500,000 \\
        \cmidrule(lr){1-4}    
            ~\textrm{FFN Dimension} & 512 & 768 & 768 \\
            ~\textrm{Activated Experts} & 8 & 8 & 8 \\
            ~\textrm{Routed Experts} & 64 & 64 & 256 \\
        \cmidrule(lr){1-4} 
            ~\textrm{Sequence Length} & 2048 & 4096 & 4096 \\
            ~\textrm{Vocab Size} & 128,256 & 128,256 & 128,256 \\
        \cmidrule(lr){1-4} 
            \#~\textrm{Dense Params} & 296 M & 479 M & 479 M \\
            \#~\textrm{Sparse Params} & 806 M  & 2.72 B & 10.88 B \\
            \#~\textrm{Activated Params} & 397 M  & 823 M & 823 M \\
            \#~\textrm{Total Params} & 1.10 B & 3.20 B & 11.36 B \\
        \bottomrule
    \end{tabular}
  }
  \caption{Hyperparameters of model architectures.}
  \label{tab:mdhyper}
\end{table}

\subsection{Optimizer Setup Details}
Table~\ref{tab:pthyper_1b} presents the hyperparameters used for pretraining the 1B MoE model with AdamW.
This parameter set was identified through our hyperparameter search over key configurations.
For pretraining 1B models with other optimizers, we simply align their update RMS with AdamW. 
This ensures fair convergence comparisons and avoid the cost for extensive hyperparameter search.

For Muon, this alignment equates to scaling the learning rate of Muon-optimized parameters by $0.2 \times \sqrt{\max(d_{in}, d_{out})}$ \cite{k2}. 
More details regarding our Muon implementation are available in the code.

For Hyperball Optimization, we fix the Frobenius norm of the weight matrix $\mW \in \mathbb{R}^{d_{in} \times d_{out}}$ at $\sqrt{d_{out}}$. 
To align the update RMS, this translates to a learning rate scaler of $0.2 \times \sqrt{d_{in}}$. 
We substitute the $d_{in}$ value from a 1B model into this formula, using the resulting value as a scale-invariant constant across all model scales.

\section{Evaluation Setup}
\label{apx:eval}
We perform all downstream task evaluations using OLMES \citep{olmes}. During pretraining, we evaluate the model on 9 core tasks—ARC-Easy \cite{arc}, ARC-Challenge \cite{arc}, MMLU \cite{mmlu}, CommonsenseQA \cite{csqa}, SocialIQA \cite{siqa}, HellaSwag \cite{hella}, WinoGrande \cite{wino}, PIQA \cite{piqa}, and SciQ \cite{sciq}—to quickly assess the fundamental capabilities of these checkpoints. Unless otherwise specified, we follow \cite{olmo3} and evaluate on a benchmark consisting of 25 multiple-choice tasks; a complete list of these tasks is provided in Table~\ref{tab:full}. To save space, we do not report the detailed scores for these 25 tasks unless explicitly specified.

\begin{table}[h]
  \centering
  \small

  \resizebox{0.66\columnwidth}{!}{
    \begin{tabular}{lc}
        \toprule
            \textbf{Configuration} & \textbf{1B} \\
        \midrule
        Optimizer & AdamW \\
        $(\beta_1, \beta_2)$ & $(0.9, 0.95)$ \\
        Peak LR & 1.0E-03 \\
        Minimum LR & 0 \\
        LR scheduler & cosine \\
        Warmup steps & 1000 \\
        Weight decay & 0.1 \\
        Gradient clip & 1.0 \\
        \cmidrule(lr){1-2}
        LBL weight & 0.01 \\
        \bottomrule
    \end{tabular}
  }
  \caption{Pretraining hyperparameters (1B AdamW).}
  \label{tab:pthyper_1b}
\end{table}

\begin{table*}[htbp]
    \small
    \resizebox{0.98\textwidth}{!}{
        \begin{tabular}{l|cc|cc|cc|cc} 
            \toprule
            \multirow{2}{*}{} & \multicolumn{2}{c}{AdamW} & \multicolumn{2}{c}{AdamH} & \multicolumn{2}{c}{Muon} & \multicolumn{2}{c}{MuonH} \\
            \cmidrule(lr{1pt}){2-3} \cmidrule(lr{1pt}){4-5} \cmidrule(lr{1pt}){6-7} \cmidrule(lr{1pt}){8-9}
            & \multicolumn{2}{c|}{MoE \; \textrm{\textit{w.} \name }} & \multicolumn{2}{c|}{MoE \; \textrm{\textit{w.} \name }} & \multicolumn{2}{c|}{MoE \; \textrm{\textit{w.} \name }} & \multicolumn{2}{c}{MoE \; \textrm{\textit{w.} \name }} \\
            \midrule
            Arc-Easy~\cite{arc} & 59.64 & 64.05 & 64.09 & 66.88 & 60.65 & 62.16 & 61.82 & 65.36\\
            Arc-Challenge~\cite{arc} & 32.51 & 36.26 & 35.15 & 37.37  & 34.90 & 35.75 & 35.24 & 33.62 \\
            MMLU-Stem~\cite{mmlu} & 28.13 & 27.80 & 28.19 & 28.66 & 28.26 & 27.60 & 27.87 & 28.33 \\
            MMLU-Humanities~\cite{mmlu} & 34.38 & 34.68 & 35.03 & 35.55 & 33.77 & 34.38 & 33.96 & 34.51 \\
            MMLU-Social Science~\cite{mmlu} & 29.12 & 29.12 & 28.20 & 29.08 & 28.48 & 29.50 & 29.25 & 29.07 \\
            MMLU-Other~\cite{mmlu} & 33.41 & 35.84 & 35.47 & 37.05 & 35.35 & 35.84 &  33.47 & 36.64 \\
            CSQA~\cite{csqa} & 38.25 & 48.73 & 38.00 & 48.08 & 44.14 & 45.37 & 42.83 & 46.76 \\
            HellaSwag~\cite{hella} & 47.18 & 47.68 & 48.52 & 48.35 & 46.52 & 48.16 & 46.31 & 45.24  \\
            WinoGrande~\cite{wino} & 52.25 & 51.93 & 51.70 & 51.93 & 51.78 & 52.01 & 52.25 & 51.93  \\
            SocialIQA~\cite{siqa} & 41.91 & 44.58 & 42.02 & 43.86 & 42.99 & 44.11 & 41.91 &43.86  \\
            PiQA~\cite{piqa} & 69.04 & 68.82& 69.04 & 69.91 & 68.61 & 69.64 & 67.74 & 67.90\\
            CoQA~\cite{coqa} & 47.09 & 52.39  & 47.00 & 37.84 & 54.37 & 53.61 & 51.27 & 54.00 \\
            DROP~\cite{drop} & 30.79 & 27.71 & 26.83 & 29.03 & 28.53 & 28.34 & 28.72 & 28.15 \\
            Jeopardy & 46.92 & 43.82 & 50.02 & 54.51 & 49.02 & 49.64 & 52.98 & 53.36 \\
            NaturalQs~\cite{nq} & 24.31 & 25.76 & 23.71 & 25.57 & 25.31 & 23.51 & 24.21 & 26.11 \\
            SQuAD~\cite{squad} & 49.91 & 48.96 & 50.47 & 51.04 & 49.86 & 48.63 & 53.65 & 52.70 \\
            SciQ~\cite{sciq} & 70.90 & 76.70 & 74.50 & 76.80 & 73.20 & 73.60 & 72.30 & 72.60 \\
            QASPER~\cite{qasper} & 67.40 & 68.02 & 58.93 & 67.40 & 62.07 & 68.03 & 61.44 & 67.40 \\
            DBQA~\cite{labbench} & 27.12 & 23.84 & 26.15 & 25.38 & 24.81 & 26.54 & 23.85 & 26.35 \\
            ProtocalQA~\cite{labbench} & 25.00 & 27.78 & 26.85 & 27.78 & 30.56 & 27.78 & 26.85 & 29.63 \\
            Lambada~\cite{lambada} & 39.34 & 39.39 & 41.57 & 40.15 & 39.76 & 39.24 & 38.95 & 40.99 \\
            MedMCQA~\cite{medmcqa} & 27.16 & 28.26 & 26.99 & 28.66 & 27.32 & 28.16 & 28.42 & 28.14\\
            MedQA~\cite{medqa} & 23.17 & 23.80 & 24.04 & 24.90 & 24.90 & 23.57 & 21.84 & 23.96 \\
            SciRIFF~\cite{sciriff} & 72.44 & 72.44 & 71.93 & 72.82 & 71.11 & 71.87 & 72.44 & 72.00 \\
            Basic Skills~\cite{olmo3} & 39.23 & 40.73 & 40.27 & 39.73 & 39.03 & 39.54 & 39.90 & 40.78 \\
            \midrule
            Avg RC Acc. & 42.26 & 43.93 & 42.59 & 43.93 & 43.01 & 43.55 & 42.78 & 43.98  \\
            \bottomrule
        \end{tabular}
    }
    \caption{Task-specific performance comparisons for 1B MoE with different optimizers.}
    \label{tab:full}
\end{table*}

\begin{figure}[h]
    \centering
    \includegraphics[width=0.98\linewidth]{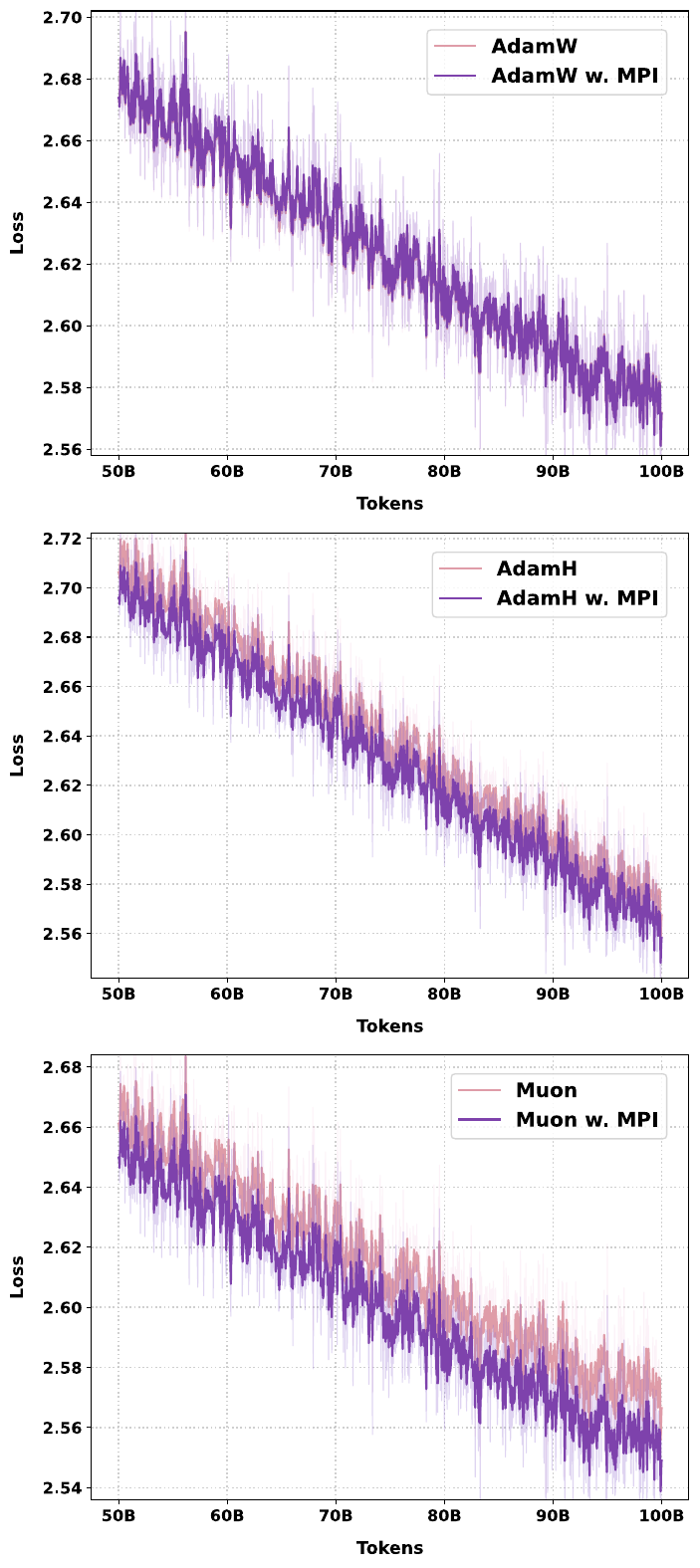}
    \caption{Pre-training loss comparison for a 1B MoE model across optimizers (AdamW, AdamH, Muon). MoE with \name\ achieves a convergence advantages over all alternative setups.}
    \label{fig:1b-loss-all}
\end{figure}

\end{document}